\documentclass[11pt]{article}
\usepackage[english]{babel}
\usepackage[utf8]{inputenc}
\usepackage{latexsym}
\usepackage{amsfonts}
\usepackage{graphicx}
\usepackage{color}
\usepackage{crayola}
\usepackage{xspace}
\usepackage{amsmath,times}

\newcommand*{\affaddr}[1]{#1} % No op here. Customize it for different styles.
\newcommand*{\affmark}[1][*]{\textsuperscript{#1}}

\title{\textbf{Exactly mergeable summaries}}
\author{
Vladimir Batagelj\affmark[1,2]  \\
ORCID: 0000-0002-0240-9446\\
\affaddr{\affmark[1]Institute of Mathematics, Physics and Mechanics,\\ Jadranska 19, 1000 Ljubljana, Slovenia}\\
\affaddr{\affmark[2]University of Primorska, Andrej Marušič Institute,\\ 6000 Koper, Slovenia} %\\ 
}

\newcommand{\clock}{\count254=\time \divide\count254 by 60
 \count255=\count254 \multiply\count255 by -60
 \advance\count255 by \time
 \ifnum\count254<10 0\fi\number\count254\,:\,%
 \ifnum\count255<10 0\fi\number\count255}

\newcommand{\keyw}[1]{\textcolor{red}{\emph{#1}}}

\newcommand{\RR}{\Bbb{R}}

\newcommand{\Mw}{\mathop{\raisebox{-1.5pt}{\mbox{$\Box$\kern-.55em\raisebox{2.5pt}{{\tiny $r$}}\kern2.9pt}}}}
\newcommand{\Mv}{\mathop{\raisebox{-1.5pt}{\mbox{$\Box$\kern-.55em\raisebox{2.5pt}{{\tiny $h$}}\kern2.9pt}}}}

\newcommand{\med}{\mbox{med}}

\graphicspath{{./}}

\date{}

\begin{document}

\maketitle

%******************************************************************************%******************************************************************************
\begin{abstract}
In the analysis of large/big data sets, \keyw{aggregation} (replacing values of a variable over a group by a single value) is a standard way of reducing the size (complexity) of the data. Data analysis programs provide different aggregation functions.

Recently some books dealing with the theoretical and algorithmic background of traditional aggregation functions were published. A problem with traditional aggregation is that often too much information is discarded thus reducing the precision of the obtained results.
A much better, preserving more information, summarization of original data can be achieved by representing aggregated data using selected types of complex data.   
  
In complex data analysis the measured values over a selected group $A$ are aggregated into a complex object $\Sigma(A)$ and not into a single value. Most of the aggregation functions theory does not apply directly. In our contribution, we present an attempt to start building a theoretical background of complex aggregation. 

We introduce and discuss \keyw{exactly mergeable} summaries for which it holds for merging of disjoint sets of units
 \[  \Sigma(A \cup B) = F( \Sigma(A),\Sigma(B)),\qquad   \mbox{ for } \quad A\cap B = \emptyset .\]

\end{abstract}

%******************************************************************************%******************************************************************************
%******************************************************************************
\section{Introduction}

%******************************************************************************
\subsection{Motivation}

In our program, Clamix \cite{Kejzar} for clustering symbolic data represented by discrete distributions $(n, \mathbf{p})$ where $\mathbf{p}$ is an empirical probability distribution and $n$ is the number of original data units summarized by $\mathbf{p}$. This representation has two important properties
\begin{itemize}
\item fixed space required for a description of a unit/cluster;
\item description of a union of two disjoint clusters can be obtained from their descriptions.
\end{itemize}
In this paper, we will elaborate on the second observation.

For example, let us consider the population pyramids of the world's countries. How to join the population pyramids of China and Vanuatu? \medskip

\noindent
\includegraphics[width=62mm]{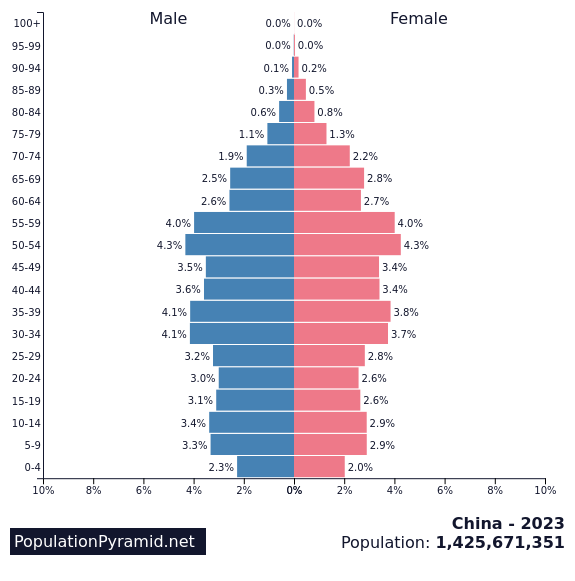} \ \ 
\includegraphics[width=62mm]{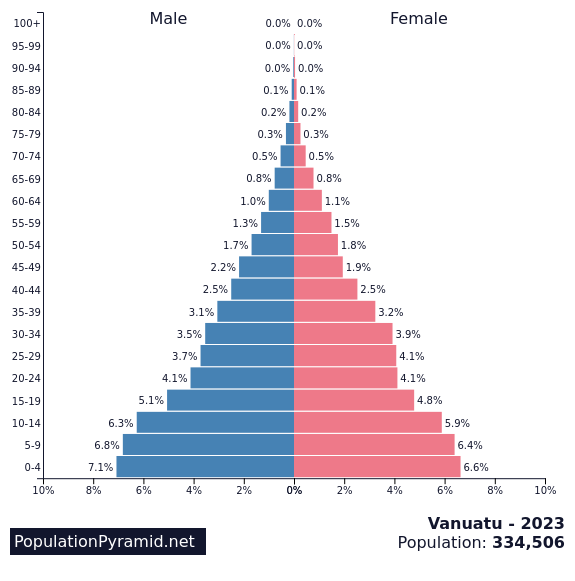}\smallskip

When comparing two countries $A$ and $B$ we compare the shapes of their probability distributions $\mathbf{p}_A$ and $\mathbf{p}_B$. But to determine the correct probability distribution of their union $A \cup B$ we need to know also the sizes $n_A$ and $n_B$ of countries $A$ and $B$
\[ (n_{A \cup B}, \mathbf{p}_{A \cup B}) = (n_A + n_B, \frac{n_A \mathbf{p}_A  + n_B \mathbf{p}_B}{n_A + n_B})\]

%******************************************************************************

\subsection{Aggregation}

In an analysis of large data sets the \keyw{aggregation} is a standard way for reducing the size (complexity) of the data. Recently some books dealing with the theoretical and algorithmic background of the traditional aggregation (replacing values of a variable over a group by a single value) were published \cite{Beliakov,Torra,Grabisch,Bustince,Halas}.

%\centerline{\includegraphics[height=38mm]{torra.jpg} \  \includegraphics[height=38mm]{beliakov.jpg} \  
%\includegraphics[height=38mm]{grabisch.jpg} \  \includegraphics[height=38mm]{bustince.jpg} }\medskip

Data analysis programs provide aggregation functions such as means (arit, geom, harm, median, modus), min, max, product, bounded sum, counting, etc. \cite{James}. Special care has to be given to variables measured in different measurement scales.

In theoretical discussion the traditional aggregation functions are usually ``normalized'' to the interval $[0,1]$ -- they take real arguments in $[0,1]^k$ and produce a value in $[0,1]$, and satisfy the conditions: $f(\mathbf{0}) = 0$,  $f(\mathbf{1}) = 1$, and monotonicity $\mathbf{x} \leq \mathbf{y} \Rightarrow f(\mathbf{x}) \leq f(\mathbf{y})$. Often, in applications, also idempotency and symmetry are required. 

The applications of traditional aggregation functions are used, besides determining a representative value for a group of measurements, mainly to combine partial criteria into a single criterion (multicriteria optimization and decision making) or to express the membership degree in combined fuzzy sets.

A problem with traditional aggregation is that often too much information is discarded thus reducing the precision of the obtained results.

 A much better, preserving more information, summarization of original data can be achieved by representing aggregated data using selected types of complex data such as symbolic objects \cite{Diday,DidayE}, compositions \cite{Aitchison}, functional data  \cite{Silverman}, etc. In the Symbolic Data Analysis (SDA) framework, much work is devoted to the summarization process, for example, the function \texttt{classic.to.sym} in RSDA \cite{Rodriguez}, and SODAS or SYR software.

%******************************************************************************
\section{Mergeable summaries}

In complex data analysis the measured values over a selected subset of units $A$ are aggregated into a complex object $\Sigma(A)$ and not into a single value. Most of the aggregation theory does not apply directly. In our contribution, we present an attempt to start building a theoretical background of complex aggregation. 

An interesting question is, which complex data types are compatible with the merging of disjoint sets of units
 \begin{equation}  \Sigma(A \cup B) = F( \Sigma(A),\Sigma(B)),\qquad   \mbox{ for } \quad A\cap B = \emptyset . \label{exact} \end{equation} 
 
Selecting a name for this kind of summary we were inclined towards the term \emph{hierarchical} or \emph{mergeable summary}. Searching on Google we learned that the term \emph{mergeable summary} was already proposed and elaborated by \cite{mergsum}.
They enable parallelization in big data algorithms and stream processing.
The summarization in big data is not deterministic and allows some errors.
A summary is \keyw{mergeable} if the error and space (size of the summary) do not increase after the merge.

In this paper, we will discuss exactly mergeable summaries ``without errors''.

%******************************************************************************
\subsection{Exactly mergeable summaries}

A summary $\Sigma(A)$ is an \keyw{exactly mergeable summary} if and only if it requires a fixed space of small size and satisfies the relation (\ref{exact}).

We can consider merging as a partially defined binary operation $ \Sigma(A) * \Sigma(B) =  F( \Sigma(A),\Sigma(B))$. For mutually disjoint subsets $A$, $B$, and $C$ we have
\[ \Sigma(A) * \Sigma(B) = \Sigma(B) * \Sigma(A) \]
\[ \Sigma(A) * (\Sigma(B) * \Sigma(C))  = (\Sigma(B) * \Sigma(A)) * \Sigma(C) \]

\subsubsection{Simple examples}

We assume that a numerical variable $v : U \to \RR$ is measured on the set of units $U$ and that $A, B \subseteq U$ and $A \cap B = \emptyset$.

Let $\mbox{sort}_A(v)$ be a list of values of the variable $v$ on the set of units $A$ ordered in decreasing order. We define $\mbox{1st}_A(v) = \mbox{sort}_A(v)[1]$
and  $\mbox{2nd}_A(v) = \mbox{sort}_A(v)[2]$.

It is easy to check that the following summaries are exactly mergeable:
\begin{enumerate}

\item  $\Sigma(A) = |A| = n_A$\\
          $\Sigma(A \cup B) = \Sigma(A)+\Sigma(B)$
\item  $\Sigma(A) = \min_{X \in A} v(X)$\\
          $\Sigma(A \cup B) = \min(\Sigma(A),\Sigma(B))$
\item  $\Sigma(A) = \max_{X \in A} v(X)$\\
          $\Sigma(A \cup B) = \max(\Sigma(A),\Sigma(B))$
\item  $\Sigma(A) = (\mbox{1st}_A(v), \mbox{2nd}_A(v))$\\
          $\Sigma(A \cup B) =  (\mbox{1st}_L(v), \mbox{2nd}_L(v))$,\\ where $L = \{  \mbox{1st}_A(v), \mbox{2nd}_A(v),  \mbox{1st}_B(v), \mbox{2nd}_B(v) \}$\\
This example can be generalized to $\Sigma(A) = \mbox{Top-}k_A(v)$.
\item  $\Sigma(A) = (n_A, \mu_A)$, \quad $\mu_A = \frac{1}{n_A}\sum_{X \in A} v(X)$\\
          $\Sigma(A \cup B) = (n_A + n_B,\frac{ n_A \mu_A + n_B \mu_B}{n_A + n_B})$
\item  $\Sigma(A) = \sum_{X \in A} v(X)$\\
          $\Sigma(A \cup B) = \Sigma(A)+\Sigma(B)$

\end{enumerate}

%******************************************************************************

\subsubsection{Moments}
% https://physics.stackexchange.com/questions/452781/how-to-calculate-the-error-when-multiple-measurements-are-made-each-of-which-ha

The distribution of values of variable $v$ on the set of units $A$ is often summarized by its average  $\mu_A$ and its standard deviation $\sigma_A$. It would be better to represent it as $\Sigma(A) = (n_A,\mu_A,\sigma_A)$, where $n_A$ is the number of units in $A$. 

Then the distribution of additional values of variable $v$ on the set of units $B$, $ A\cap B = \emptyset$, is summarized by $\Sigma(B) = (n_B,\mu_B,\sigma_B)$ and can be combined into a summary of the distribution on the set $C = A \cup B$,  $\Sigma(C) = (n_C,\mu_C,\sigma_C)$  determined by $\Sigma(A)$ and $\Sigma(B)$ as follows
\[ n_C = n_{A\cup B} = n_A + n_B \]
\[ \mu_C = \mu_{A\cup B} = \frac{n_A \mu_A + n_B \mu_B}{n_C}\]
\[ \sigma_C = \sigma_{A\cup B} = \sqrt{\frac{S_C}{n_C} - \mu_C^2}\]
where $S_C = S_A + S_B$ and $S_X = n_X (\sigma_X^2 + \mu_X^2)$.  $\Sigma(A)$ is an exactly mergeable summary.

This result can be extended to higher moments.
% In a similar way we discuss some other representations used in the construction of complex objects by aggregation.

%******************************************************************************

\subsubsection{Set membership count}

Counting the number of units from $C$ in $A$
\[ n(A;C) = |A \cap C| \]
is an exactly mergeable summary.

\noindent\textbf{Proof:}
\[ n(A\cup B; C) = | (A\cup B) \cap C|  = | (A \cap C_) \cup (B \cap C) | = \]
\[  =  | A\cap C| + |B\cap C| - | A \cap B \cap C |  = n(A;C) + n(B;C) \quad \quad \Box\]

%******************************************************************************

\subsubsection{Combining exactly mergeable summaries}

Let $\Sigma_1$ and  $\Sigma_2$ be exactly mergeable summaries. Then also their \keyw{composition}
\[ \Sigma_1 \oplus \Sigma_2  (A) = ( \Sigma_1 (A),  \Sigma_2(A) ) \]
is an exactly mergeable summary.

\noindent\textbf{Proof:}
$ \displaystyle \Sigma_1 \oplus \Sigma_2  (A \cup B) = ( \Sigma_1 (A \cup B),  \Sigma_2(A \cup B) ) = $
\[ = (  F_1( \Sigma_1(A),\Sigma_1(B)),  F_2( \Sigma_2(A),\Sigma_2(B)) )  \quad \quad \Box \]
 
%******************************************************************************

Since min and max are mergeable summaries also their composition -- the \keyw{interval summary} of the variable $v$ on the set of units $A$ 
\[  \Sigma(A) = [ \min_{X \in A} v(X), \max_{X \in A} v(X)] \]
is an exactly mergeable summary.
 Let $\Sigma(A) = [m_A, M_A]$ and $\Sigma(B) = [m_B, M_B]$ then
\[ \Sigma(A\cup B) = [ \min_{X \in A\cup B} v(X), \max_{X \in A\cup B} v(X)]= [\min(m_A,m_B), \max(M_A,M_B)]\]

Let $K = \{k_1,k_2,\ldots,k_s\}$ be a finite set of categories and $v : U \to K$ a categorical (nominal) variable on the set of units $U$. The summary
\[ \Sigma(A) = \{ (k, n(A,C(k))) : k \in K \} \quad 
    \mbox{where} \quad\ C(k) = \{ X : v(X) = k \} \]
is called a \keyw{bar chart}.

Let $v : U \to \RR$ be an ordinal variable and $\mathbf{B} = (B_1,B_2,\ldots,B_r)$ an ordered partition (set of bins) of $v(A)$.  The summary
\[ \Sigma(A) = [ (B, n(A,C(B))) : B \in \mathbf{B} ] \quad  
    \mbox{where} \quad\ C(B) = \{ X : v(X) \in B \} \]
is called a \keyw{histogram}. 

A histogram (and also a bar chart) is essentially a frequency distribution $\mathbf{f}$ over a given set of bins $\mathbf{B}$ (categories $K$). It can be equivalently representent by a pair $(n,\mathbf{p})$ where $n = \sum_i f_i$ is the size of the set $A$ and $\mathbf{p} = \frac{1}{n} \mathbf{f}$ is the corresponding probabilty distribution.

Therefore, since set membership counts are exactly mergeable, the bar charts
and histograms are exactly mergeable summaries.

%******************************************************************************

\subsubsection{Proving that a summary is not  exactly mergeable}

If for a summary $\Sigma$ exist sets $A_1$, $B_1$, $A_2$, $B_2$ such that $A_1\cap B_1 = \emptyset$,
 $A_2\cap B_2 = \emptyset$, $\Sigma(A_1) = \Sigma(A_2)$,   $\Sigma(B_1) = \Sigma(B_2)$,  and
 $\Sigma(A_1\cup B_1) \ne \Sigma(A_2\cup B_2)$ then $\Sigma$ \textbf{is not}  exactly mergeable.
 
\noindent\textbf{Proof:} Assume that  $\Sigma$ is  exactly mergeable. Then
 \[ \Sigma(A_1\cup B_1) = F( \Sigma(A_1),\Sigma(B_1)) = F( \Sigma(A_2),\Sigma(B_2)) =  \Sigma(A_2\cup B_2)\]
-- a contradiction.

%******************************************************************************

\noindent\textbf{Example 1.} Median is not  exactly mergeable  summary
%\small
\[ \med_A(v) = \mbox{sort}_A(v)[ \left \lceil \frac{n_A}{2}\right \rceil ]  \]

\begin{tabular}{llcll} 
$v(A_1) = [ 3, 4, 1] $&$ \med_{A_1}(v) = 3$ && $v(A_2) = [ 3, 8 ] $&$ \med_{A_2}(v) = 3$ \\
$v(B_1) = [ 9, 6] $&$ \med_{B_1}(v) = 6$ && $v(B_2) = [ 6, 2, 7] $&$ \med_{B_2}(v) = 6$ \\
&$\med_{A_1 \cup B_1}(v) = 4$ &&& $\med_{A_2 \cup B_2}(v) = 6$ \\ 
\end{tabular}
%\medskip

%******************************************************************************
\noindent\textbf{Example 2.} 2nd is not exactly mergeable summary

%\[ \mbox{2nd}_A(v) = \mbox{sort}_A(v)[ 2 ]  \]

\begin{tabular}{llcll}
$v(A_1) = [ 1, 3, 5] $&$ \mbox{2nd}_{A_1}(v) = 3$ && $v(A_2) = [ 3, 3, 6 ] $&$ \mbox{2nd}_{A_2}(v) = 3$ \\
$v(B_1) = [ 2, 5, 6] $&$ \mbox{2nd}_{B_1}(v) = 5$ && $v(B_2) = [ 4, 5, 7] $&$ \mbox{2nd}_{B_2}(v) = 5$ \\
& $\mbox{2nd}_{A_1 \cup B_1}(v) = 2$ &&& $\mbox{2nd}_{A_2 \cup B_2}(v) = 3$ 
\end{tabular}

%******************************************************************************
\section{Conclusions}
%******************************************************************************

%******************************************************************************

In this paper, we introduced the notion of exactly mergeable summaries. We showed that the summaries interval, max, min, top-$k$, bar chart, and histogram (used in SDA) are exactly mergeable.  Adding the size of the set of units makes some summaries, such as moments and discrete probability distribution, exactly mergeable.

%******************************************************************************
\section*{Acknowledgments}

This paper is an elaboration of ideas presented at 
the 7th Workshop on Symbolic Data Analysis, SDA 2018, held in Viana do Castelo, Portugal,
 18 -- 20 October 2018. It was presented at NTTS2023 -- Conference on New Techniques and Technologies for Statistics, 6 -- 10 March 2023 (Bruxelles, Belgium).

This work is supported in part by the Slovenian Research Agency
 (research program P1-0294 and research projects J5-2557, J1-2481 and J5-4596),
and prepared within the framework of the COST action CA21163 (HiTEc).

%\section*{References}

%******************************************************************************
\end{document}